%% file: neurips_2026.tex
\documentclass{article}

\usepackage[preprint]{neurips_2026}
\usepackage{wrapfig}
\usepackage{pifont}

\usepackage[utf8]{inputenc} 
\usepackage[T1]{fontenc}    
\usepackage{hyperref}       
\usepackage{url}            
\usepackage{booktabs}       
\usepackage{amsfonts}       
\usepackage{nicefrac}       
\usepackage{microtype}      
\usepackage[table]{xcolor}  
\usepackage{multirow}
\usepackage{graphicx}
\usepackage{amsmath}
\usepackage{subcaption}
\usepackage{algorithm}
\usepackage{algpseudocode}

\usepackage{amsthm}

\theoremstyle{remark}
\newtheorem{remark}{Remark}
\definecolor{starblue}{HTML}{2F5F9B}
\definecolor{algteal}{HTML}{0A8A8F}
\definecolor{eqblue}{HTML}{1E4FA3}
\newcommand{\starplug}{\textbf{\textcolor{starblue}{~+~\textsc{STARS}}}}
\newcommand{\pairrule}{\cmidrule[0.06em](lr){2-8}}
\algrenewcommand\algorithmicrequire{\textbf{Require:}}
\algrenewcommand\algorithmicensure{\textbf{Ensure:}}
\algrenewcommand\algorithmiccomment[1]{\hfill{\color{algteal}\#~#1}}
\newcommand{\algstage}[1]{\Statex{\color{algteal}\#~#1}}
\newcommand{\eqcomment}[1]{\Comment{Eq.~\textcolor{eqblue}{\ref{#1}}}}

\usepackage[capitalize,noabbrev]{cleveref}

\usepackage{tcolorbox}
\newtcolorbox{chronobox}{
  colframe=black!65,
  colback=gray!5,
  boxrule=0.9pt,
  arc=4mm,
  left=10pt,
  right=10pt,
  top=7pt,
  bottom=7pt,
  boxsep=0pt
}

\title{STARS: Spike Tail-Aware Relational Synthesis for ANN-to-SNN Data-Free Knowledge Distillation}

\author{%
\parbox{\textwidth}{\centering
\textbf{Shuhan Ye}$^{1}$,
\textbf{Yi Yu}$^{2}$,
\textbf{Qixin Zhang}$^{3}$,
\textbf{Hui Lu}$^{1}$,
\textbf{Jiaming He}$^{1}$,
\textbf{Qinggang Zhang}$^{2}$,
\textbf{Li Shen}$^{4}$,
\textbf{Xudong Jiang}$^{1}$\\[1mm]
{\normalfont
\begin{tabular}{@{}c@{\quad}c@{\quad}c@{}}
\multicolumn{3}{c}{$^{1}$Nanyang Technological University}\\
$^{2}$Jilin University &
$^{3}$Wuhan University &
$^{4}$Sun Yat-sen University
\end{tabular}\\
\texttt{SHUHAN006@e.ntu.edu.sg}
}
}
}
\begin{document}

\maketitle

\begin{abstract}
SNNs promise energy-efficient and low-latency inference, but their performance still trails that of ANNs. ANN-to-SNN knowledge distillation helps narrow this gap, yet the original training data are often unavailable in practical deployment settings. Existing data-free knowledge distillation (DFKD) methods synthesize surrogate data by matching teacher-side priors, especially BN statistics, but these ANN-oriented constraints mainly regularize mean and variance and therefore remain under-constrained for SNN students whose responses depend on threshold-crossing dynamics. In this paper, we propose Spike Tail-Aware Relational Synthesis (STARS), a plug-and-play method for ANN-to-SNN DFKD that augments standard BN-guided synthesis with two complementary objectives: Relational Consistency Alignment, which preserves cross-sample relational consistency between teacher and student, and Tail-Aware Regularization, which regularizes threshold-relevant tail probabilities through soft exceedance over teacher-derived thresholds. Together, these objectives generate synthetic batches that remain teacher-valid while becoming more informative for SNN students. Experiments on CIFAR-10, CIFAR-100, and Tiny-ImageNet across multiple ANN-SNN pairs show that our method consistently improves conventional DFKD baselines and even surpasses several KD methods, with gains of up to 4.6\% on CIFAR-10 and 6.7\% on CIFAR-100, highlighting the importance of complementing BN matching with relational and tail-aware constraints in SNN-oriented DFKD.
\end{abstract}

\section{Introduction}
\label{sec:introduction}

Spiking neural networks (SNNs) have been widely regarded as the third generation of neural networks because they compute with discrete spikes and explicitly model temporal neural dynamics ~\citep{maass1997,surrogate,roy2019towards,wu2018spatio,zheng2021going}. Their event-driven communication paradigm offers an attractive route to energy-efficient and low-latency inference, especially for edge intelligence and neuromorphic hardware ~\citep{roy2019towards,davies2018loihi,fang2021incorporating,deng2021optimal,qian2025ucf}. Recent progress in direct training and surrogate-gradient learning has further improved the practicality of modern SNNs ~\citep{surrogate,li2021differentiable,yao2022glif,zhou2023spikformer,wu2021progressive,yang2022training,wei2023temporal,guo2022recdis}. However, directly trained SNNs still often suffer from limited optimization stability and suboptimal task accuracy in practical applications, which has motivated ANN-to-SNN knowledge distillation~\citep{hinton2015distilling,romero2015fitnets,zagoruyko2017paying,park2019relational,tian2020contrastive} as a way to transfer richer supervision from strong ANN teachers while preserving efficient SNN inference~\citep{guo2023joint,bkdsnn,ckd,tser,efficientannguided}.

In many practical settings, however, the original training data are unavailable at distillation time because of privacy, copyright, transmission, or storage constraints~\citep{yoo2019knowledge,fang2019data,binici2022robust}. This setting is especially relevant in edge deployment scenarios for SNNs, where pretrained models may remain accessible while the original data can no longer be retained locally or transmitted across devices. Data-free knowledge distillation (DFKD) addresses this scenario by synthesizing surrogate samples from a pretrained teacher and then using them for transfer ~\citep{micaelli2019zero,nayak2019zero,dafl,yin2020dreaming,cmi,nayer}. As a result, DFKD provides an attractive route for ANN-to-SNN transfer when real data are inaccessible.

\begin{figure}
    \centering
    \includegraphics[width=1\linewidth]{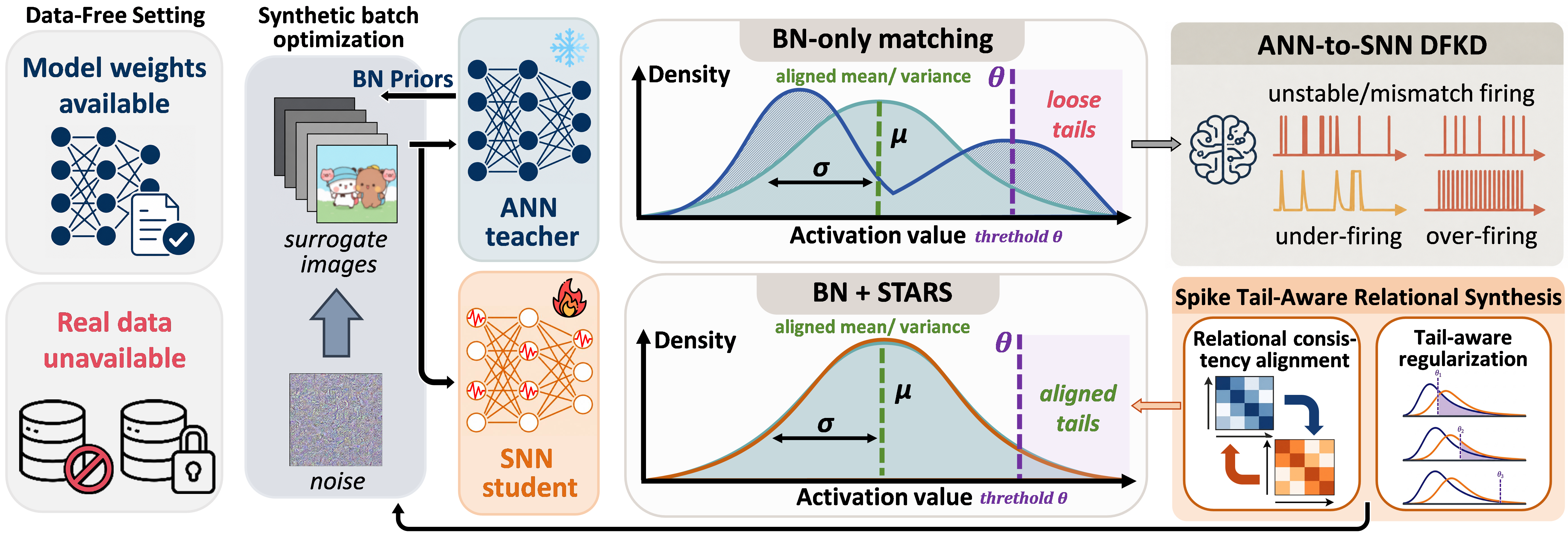}
    \vspace{-4mm}
    \caption{Overview of our proposed method STARS. BN-guided synthesis provides teacher-valid surrogate data but only constrains low-order activation statistics, leaving threshold-relevant tails ambiguous for SNN students. STARS complements BN matching with relational consistency alignment and tail-aware regularization to synthesize data that better matches SNN firing behavior.}
    \label{fig:method}
\vspace{-4mm}
\end{figure}

Previous work has shown that the quality of synthetic data depends critically on the priors used during synthesis. Early methods explored zero-shot and generator-based formulations ~\citep{micaelli2019zero,dafl}, while later works demonstrated the importance of internal teacher statistics, especially Batch Normalization (BN) statistics, for reconstructing more faithful surrogate data ~\citep{yin2020dreaming}. Subsequent methods further improved data synthesis with stronger diversity, contrastive, and optimization priors ~\citep{cmi,nayer,agnosticdfml,apidfml,free,taskgroupings,robustdfml}. Modern DFKD methods therefore rely heavily on teacher-side priors, with BN matching serving as one of the most effective and widely adopted anchors for sample synthesis.

However, directly applying existing DFKD to ANN-to-SNN distillation may be suboptimal. In conventional DFKD, both teacher and student are ANNs, so teacher-driven synthesis with BN matching is typically sufficient to generate samples compatible with the ANN student. In ANN-to-SNN distillation, however, the synthetic data remain optimized toward ANN teacher priors, whereas the SNN student follows distinct dynamics induced by integration, thresholding, and reset. This asymmetry can introduce a mismatch: samples satisfying ANN BN constraints may still elicit inconsistent responses in the SNN student. In this work, we analyze this gap and show that ANN-to-SNN DFKD should go beyond teacher-side BN statistics to additionally constrain cross-sample relations and threshold-relevant tail behavior.

To address this mismatch, we propose \textbf{S}pike \textbf{T}ail-\textbf{A}ware \textbf{R}elational \textbf{S}ynthesis (\textbf{STARS}), a plug-and-play framework for ANN-to-SNN DFKD. As shown in Figure~\ref{fig:method}, STARS combines \emph{Relational Consistency Alignment} (RCA), which preserves cross-sample relational consistency between teacher and student, with \emph{Tail-Aware Regularization} (TAR), which regularizes soft exceedance behavior over teacher-derived thresholds relevant to LIF firing. Together, they complement BN matching and make the synthesized data more informative for SNN training.

Our main contributions are: 
\begin{itemize}
    \item To the best of our knowledge, this is the first work to study ANN-to-SNN data-free knowledge distillation, and we analyze the mismatch induced by ANN-biased synthesis in SNN students, providing insights that may benefit future SNN DFKD research. 
    \item We propose STARS, a plug-and-play method that augments standard BN-guided synthesis with RCA and TAR to preserve cross-sample relational consistency and constrain threshold-relevant tail statistics for SNN training.
    \item We plug STARS into multiple conventional DFKD methods and evaluate it on CIFAR-10, CIFAR-100, and Tiny-ImageNet, where STARS consistently delivers plug-and-play performance gains across diverse ANN-SNN teacher-student pairs.
\end{itemize}

\vspace{-2mm}
\section{Preliminary and Related Work}
\label{sec:prelim}
\vspace{-1mm}

\textbf{Spiking neuron dynamics.}
Spiking neural networks (SNNs) replace continuous activations with binary spikes and perform computation through membrane integration, threshold-triggered firing and reset~\citep{surrogate}.
Following the widely used discrete-time formulation of the leaky integrate-and-fire (LIF) neuron~\citep{gerstner2014neuronal}, we denote by $X[t]$, $H[t]$, $S[t]$, and $V[t]$ the input current, the sub-threshold membrane potential, the output spike, and the post-reset membrane potential at time step $t$, respectively.
The neuron dynamics is written as
\begin{equation}\small
H[t] \!=\! V[t-1] + \frac{1}{\tau}(X[t] - (V[t-1]-V_{\mathrm{r}})), ~S[t] \!=\! \Theta(H[t]-V_{\mathrm{th}}), ~V[t] \!=\! H[t](1-S[t]) + V_{\mathrm{r}}S[t],
\label{eq:lif}
\end{equation}
where $\tau$ is the membrane time constant, $V_{\mathrm{th}}$ is the firing threshold, $V_{\mathrm{r}}$ is the reset potential, and $\Theta(\cdot)$ is the Heaviside step function.
As shown in Eq.~\eqref{eq:lif}, spiking neuron dynamics differ fundamentally from those of ANNs, as their outputs depend on temporally evolving spiking states shaped by membrane integration, thresholding, and reset, rather than static activation mappings.

\textbf{Knowledge distillation for SNNs.}
Knowledge distillation (KD)~\citep{hinton2015distilling,romero2015fitnets,zagoruyko2017paying,park2019relational,tian2020contrastive} has demonstrated strong effectiveness for SNN training~\citep{guo2023joint,bkdsnn,tser,efficientannguided}, where knowledge from a powerful teacher model is transferred to an SNN student to improve its performance.
For a mini-batch $\mathcal{B}$, a student model $f_S$ is typically trained to mimic a teacher model $f_T$ by matching softened predictions:
\begin{equation}\small
\mathcal{L}_{\mathrm{KD}}
=
\frac{1}{|\mathcal{B}|}\sum_{x\in\mathcal{B}}
\left[
\mathrm{KL}\!\left(
\sigma\!\left(f_T(x)/\gamma\right)
\mid
\sigma\!\left(f_S(x)/\gamma\right)
\right)
\right],
\label{eq:kd}
\end{equation}
where $\sigma(\cdot)$ is the softmax and $\gamma$ is the temperature.
Since existing SNNs still lag behind their ANN counterparts in accuracy, more studies have explored distilling knowledge from ANN teachers to SNNs, aiming to improve SNN performance while preserving their energy-efficiency advantage.
~\cite{bkdsnn} introduces blurred feature distillation to better mimic ANN representations,
~\cite{tser} performs temporally separated logit distillation with entropy regularization, and ~\cite{efficientannguided} aligns rate-based SNN features with ANN features via hybrid block-wise replacement. 

However, many practical scenarios do not permit access to the original training data, due to privacy, storage, or deployment constraints.
This challenge is particularly relevant for SNNs, which are often deployed in resource-constrained and edge settings, motivating the study of transferring knowledge to students without access to the original training data~\citep{roy2019towards,yin2020dreaming}.

\textbf{Data-free knowledge distillation (DFKD).}
To address the absence of original training data caused by privacy, storage, and deployment constraints, researchers have developed DFKD, which synthesizes surrogate data from a pretrained teacher to achieve knowledge transfer without access to the real training set.
~\cite{dafl} synthesizes data by exploiting teacher outputs and class-discriminative priors, ~\cite{yin2020dreaming} reconstructs inputs by matching the internal batch-normalization (BN) statistics stored in the teacher, ~\cite{cmi} improves sample diversity and transferability through contrastive model inversion on top of teacher-side priors, and ~\cite{nayer} further enhances synthesis quality within the BN-guided DFKD framework~\citep{agnosticdfml,apidfml,free,taskgroupings,robustdfml}.
These studies suggest that teacher-side BN statistics serve as an important anchor in DFKD, constraining the synthesized data to remain semantically consistent with the original training distribution and thereby providing effective supervision for subsequent distillation. 

Specifically, for the $l$-th BN layer of the teacher network, let $\hat{\mu}_l$ and $\hat{\sigma}_l^2$ be the running mean and variance estimated from the real training data.
For a synthetic mini-batch $\mathcal{B}=\{x_i\}_{i=1}^{B}$, let $\mu_l(\mathcal{B})$ and $\sigma_l^2(\mathcal{B})$ denote the empirical mean and variance of the corresponding pre-BN activations.
A standard BN-matching regularization is defined as
\begin{equation}\small
\mathcal{L}_{\mathrm{BN}}(\mathcal{B})
=
\sum_{l}
\Big(
\|\mu_l(\mathcal{B})-\hat{\mu}_l\|_2^2
+
\|\sigma_l^2(\mathcal{B})-\hat{\sigma}_l^2\|_2^2
\Big).
\label{eq:lbn}
\end{equation}
The synthetic batch is then optimized by combining BN matching with teacher-oriented objectives such as class confidence and image regularization:
\begin{equation}\small
\mathcal{B}^{\star}
=
\arg\min_{\mathcal{B}}
\Big[
\mathcal{L}_{\mathrm{cls}}(\mathcal{B}; f_T)
+
\lambda_{\mathrm{BN}}\mathcal{L}_{\mathrm{BN}}(\mathcal{B})
+
\lambda_{\mathrm{reg}}\mathcal{R}(\mathcal{B})
\Big].
\label{eq:dfkd}
\end{equation}
In this way, DFKD synthesizes surrogate samples by aligning teacher-side feature statistics with those induced by the inaccessible real data. Such priors are naturally suited to conventional ANN-to-ANN DFKD, where both teacher and student process continuous activations. For ANN-to-SNN distillation, however, the target statistics in Eq.~\eqref{eq:lbn}, \textit{i.e.,} $\hat{\mu}_l$ and $\hat{\sigma}_l^2$, are from continuous ANN activations, whereas the same synthetic batch induces student responses shaped by the LIF dynamics in Eq.~\eqref{eq:lif}. This discrepancy motivates a natural question: 
\begin{chronobox}
\centering
\textit{Are BN moments sufficient to determine the responses that matter for a SNN student?} 
\end{chronobox}

\section{Analysis: BN Priors are Under-Constrained for SNNs}
\label{sec:analysis}
In this section, we answer the above question with the following analysis. We consider the most common direct coding setting for SNNs, where a static image is presented identically at each time step. Under this setting, we analyze using a BN-constrained ANN layer together with its corresponding SNN layer. This local analysis isolates the mechanism behind the mismatch. In deep SNNs, the same effect can propagate through SNN layers. Let $a^l=W^l z^{l-1}+b^l$ denote the pre-BN activation of the $l$-th ANN teacher layer. For simplicity, we omit the layer index and write $a=wz+b$. Corresponding to the BN statistics $\mu_l(\mathcal{B})$ and $\sigma_l^2(\mathcal{B})$ in Eq.~\eqref{eq:lbn}, we write their distribution-level counterparts under the induced synthetic distribution $P$ as
\begin{equation}\small
\mu_A(P)=\mathbb{E}_{a\sim P}[a], \qquad
\sigma_A^2(P)=\mathrm{Var}_{a\sim P}(a).
\label{eq:ann_bn_stats_analysis}
\end{equation}
Therefore, any two distributions with the same $(\mu_A,\sigma_A^2)$ are equivalent from the teacher's BN perspective. However, when transferring the teacher's knowledge to an SNN student, this equivalence no longer holds, because the student does not consume $a$ as a continuous activation. Using Eq.~\eqref{eq:lif} and setting $V_{\mathrm{r}}=0$ for simplicity, we obtain the subthreshold dynamics
\begin{equation}\small
H[t] \approx \Big(1-\frac{1}{\tau}\Big) H[t-1] + \frac{1}{\tau} a,
\quad H[0]=0
\quad\Rightarrow\quad
H[t]=\beta_t a,\quad
\beta_t := 1-\Big(1-\frac{1}{\tau}\Big)^t.
\label{eq:subthreshold_analysis}
\end{equation}
The left-hand recursion describes the subthreshold membrane update induced by the constant input $a$. Under the subthreshold regime, no spike has been generated yet, so $S[t-1]=0$ and thus $V[t-1]=H[t-1]$. Substituting this relation together with the constant input $X[t]=a$ into Eq.~\eqref{eq:lif} gives the linear recursion
\begin{equation}\small
H[t] = H[t-1] + \frac{1}{\tau}(a-H[t-1])
= \Big(1-\frac{1}{\tau}\Big)H[t-1] + \frac{1}{\tau}a,
\label{eq:subthreshold_recursion}
\end{equation}
The right-hand expression in Eq.~\eqref{eq:subthreshold_analysis}, namely $H[t]=\beta_t a$, is the closed-form solution of Eq.~\eqref{eq:subthreshold_recursion} with $H[0]=0$. Here, $\beta_t$ is a time-dependent integration factor that quantifies how much of the constant input $a$ has been accumulated by time step $t$. Therefore, before threshold crossing, the membrane potential  $H[t]$ is simply a scaled version of the pre-BN activation $a$, with the scaling coefficient gradually increasing over time.

As a result, threshold crossing at time step $t$ occurs if $H[t]\ge V_{\mathrm{th}}$, \textit{i.e.,} if $\beta_t a \ge V_{\mathrm{th}}$. Equivalently, the spike response can be written in terms of a time-dependent effective threshold $\theta_t$:
\begin{equation}\small
S[t]=\Theta(\beta_t a - V_{\mathrm{th}})
=\mathbf{1}(a\ge \theta_t),
\quad
\theta_t := \frac{V_{\mathrm{th}}}{\beta_t},
\label{eq:spike_response_analysis}
\end{equation}
This shows that, under the subthreshold approximation, the spike event at time step $t$ is equivalent to a threshold-crossing event on the pre-BN activation $a$, with a time-dependent effective threshold $\theta_t$. Accordingly, the average firing rate over $T$ steps is $\bar S_T(a)=\frac{1}{T}\sum_{t=1}^{T}\mathbf{1}(a\ge \theta_t)$. To connect this threshold-crossing behavior with the activation distribution induced by synthetic data, let $F_P$ denote the cumulative distribution function (CDF) of $a$ under $P$. The expected firing rate of the SNN is
\begin{equation}\small
R_T(P)
:=
\mathbb{E}_{a\sim P}\!\left[\bar S_T(a)\right]
=
\frac{1}{T}\sum_{t=1}^{T}\mathbb{P}_{a\sim P}(a\ge \theta_t)
=
\frac{1}{T}\sum_{t=1}^{T}\bigl(1-F_P(\theta_t)\bigr).
\label{eq:rate_analysis}
\end{equation}
Eq.~\eqref{eq:rate_analysis} shows that the expected SNN response is determined by the exceedance probabilities above the effective thresholds $\{\theta_t\}_{t=1}^{T}$, rather than solely by the first and second order statistics of $a$. Now consider two activation distributions $P$ and $Q$ such that
\begin{equation}\small
\mu_A(P)=\mu_A(Q), \qquad \sigma_A^2(P)=\sigma_A^2(Q).
\label{eq:moment_match_analysis}
\end{equation}
Although they are indistinguishable under BN matching, their induced SNN responses satisfy
\begin{equation}\small
R_T(P)-R_T(Q)
=
\frac{1}{T}\sum_{t=1}^{T}\bigl(F_Q(\theta_t)-F_P(\theta_t)\bigr).
\label{eq:rate_gap_analysis}
\end{equation}
Therefore, the effective SNN response is governed by the \emph{tail masses} of the activation distribution at the threshold set $\{\theta_t\}$, rather than solely by the first and second order statistics in Eq.~\eqref{eq:ann_bn_stats_analysis}. Consequently, ANN-side BN matching leaves an ambiguity from the SNN perspective: activation distributions with identical first and second order statistics may still induce different spike responses whenever they assign different probability mass above the effective thresholds, and this discrepancy may further accumulate and amplify in deep networks through the layerwise propagation of LIF dynamics. \textbf{This directly motivates constraining threshold-relevant tail behavior in ANN-to-SNN DFKD, rather than relying on BN statistics alone.} Further remarks are provided in Appendix~\ref{app:bn_ambiguity}.

\section{Methodology}
\label{sec:method}
\vspace{-1mm}
\textbf{Problem Statement.} The analysis in Sec.~\ref{sec:analysis} identifies the central difficulty of ANN-to-SNN DFKD. Although the teacher-side BN objective in Eq.~\eqref{eq:lbn} constrains the first and second order statistics of ANN activations, the SNN student responds according to threshold-crossing events induced by the LIF dynamics in Eq.~\eqref{eq:lif}. In particular, Eq.~\eqref{eq:spike_response_analysis}--\eqref{eq:rate_gap_analysis} show that two synthetic activation distributions can be indistinguishable under BN matching yet still induce different SNN responses when they place different probability mass above the effective thresholds $\{\theta_t\}$. This analysis reveals a teacher-valid but student-ambiguous regime. \emph{This ambiguity is not a peripheral artifact but an inherent challenge that must be addressed when implementing BN-prior-based ANN-to-SNN DFKD.}

\textbf{Motivation.}
Because this issue is intrinsic rather than incidental, we seek a plug-and-play solution that can be added on top of the original BN-based matching objective. The goal is to retain the original BN prior while making the synthesized batches more informative for the SNN student by constraining additional threshold-relevant information that BN matching alone does not determine. Meanwhile, since the LIF dynamics in Eq.~\eqref{eq:lif} involve thresholding and reset, directly regularizing post-spike outputs is less suitable for optimization. Accordingly, we impose the additional constraint on \emph{pre-activation} features, consistent with the analysis in Sec.~\ref{sec:analysis}, which characterizes the response in terms of the pre-threshold activation $a$ and the effective thresholds $\theta_t$.

\textbf{The Proposed Framework: STARS.}
In this paper, we design \emph{\textbf{S}pike \textbf{T}ail-\textbf{A}ware \textbf{R}elational \textbf{S}ynthesis} (STARS), a plug-and-play regularization framework for ANN-to-SNN DFKD. STARS is built on top of the original BN-guided synthesis objective and introduces two complementary components: \emph{Relational Consistency Alignment} (RCA), which preserves global semantics structure by aligning cross-sample relational consistency between teacher and student, and \emph{Tail-Aware Regularization} (TAR), which regularizes the threshold-relevant tail behavior that BN matching alone does not determine to reduce local threshold-induced ambiguity. In this way, STARS preserves the original teacher-side prior while making the synthesized batches more informative for the SNN student.

Our STARS is imposed on \emph{pre-activation} features, \textit{e.g.,} before ReLU in the ANN teacher and before LIF in the SNN student. This choice is motivated by the LIF dynamics in Eq.~\eqref{eq:lif}: due to thresholding and reset, post-spike responses are less amenable to direct optimization. In contrast, pre-activation features provide a smoother and more comparable space for regularization. This design is also consistent with the analysis in Sec.~\ref{sec:analysis}, where the student response is characterized by the pre-threshold activation and the effective thresholds. 

For each selected layer $l\in\mathcal{S}$, let $F_T^l\in\mathbb{R}^{B\times C\times N}$ denote the teacher pre-activation feature and $F_S^l\in\mathbb{R}^{T\times B\times C\times N}$ denote the corresponding student feature produced by the current synthetic batch $\mathcal{B}$, where $B$ is the batch size, $C$ is the channel dimension, $N$ is the number of spatial locations after flattening, and $T$ is the number of time steps. We obtain global features by averaging the teacher feature over spatial locations and the student feature over both time steps and spatial locations:
\vspace{-1mm}
\begin{equation}\small
z_T^l=\frac{1}{N}\sum_{n=1}^{N}F_{T,n}^l\in \mathbb{R}^{B\times C},\qquad
z_S^l=\frac{1}{TN}\sum_{t=1}^{T}\sum_{n=1}^{N}F_{S,t,n}^l\in \mathbb{R}^{B\times C}.
\label{eq:method_features}
\end{equation}
where $F_{T,n}^l$ denotes the $n$-th spatial slice of $F_T^l$, and $F_{S,t,n}^l$ denotes the slice of $F_S^l$ at time step $t$ and spatial index $n$. The resulting $z_T^l$ and $z_S^l$ form a unified batch-wise channel representation, where rows index samples and columns index channels, corresponding to the global pre-activation features of the ANN teacher and the SNN student, respectively.


\textbf{Relational Consistency Alignment (RCA).}
The first component of STARS is designed to address the under-constrained nature of conventional BN-guided synthesis in ANN-to-SNN DFKD. As shown in Eq.~\eqref{eq:rate_gap_analysis}, synthetic samples that are valid under ANN-side BN matching can still remain ambiguous from the SNN perspective. As a result, the synthesized images may inherit an ANN-oriented bias, leading to a mismatch between how they are organized in the feature spaces of the ANN teacher and the SNN student. This mismatch can accumulate across layers and progressively distort the feature structure perceived by the student. 

To mitigate this effect, we introduce a global \emph{Relational Consistency Alignment} (RCA) objective that aligns the pairwise relations among samples within each batch between teacher and student. By anchoring the batch-level relational structure seen by the student to that induced by the teacher, RCA regularizes the global organization of synthetic samples and promotes more consistent teacher-student alignment throughout the network.

Specifically, we first apply row-wise $\ell_2$ normalization to the global pre-activation features $z_T^l$ and $z_S^l$ in Eq.~\eqref{eq:method_features}, yielding $\tilde z_T^l,\tilde z_S^l\in\mathbb{R}^{B\times C}$. We then construct cosine relation matrices
\begin{equation}\small
G_T^l = \tilde z_T^l (\tilde z_T^l)^{\mathsf T},
\qquad
G_S^l = \tilde z_S^l (\tilde z_S^l)^{\mathsf T},
\label{eq:rca_gram}
\end{equation}
where each entry measures the cosine similarity between a pair of samples in the batch. Since the diagonal entries are trivially equal to one, we match only the off-diagonal relations,
\begin{equation}\small
\ell_{\mathrm{rca}}^l
=
\frac{1}{|\mathcal{P}|}
\sum_{(i,j)\in\mathcal{P}}
\left(G_{S,ij}^l - G_{T,ij}^l\right)^2,
\qquad
\mathcal{P} := \{(i,j)\mid 1\le i<j\le B\},
\label{eq:rca_layer}
\end{equation}
and aggregate them over layers as
\begin{equation}\small
\mathcal{L}_{\mathrm{RCA}}
=
\frac{1}{\sum_{l\in\mathcal{S}} w_l}
\sum_{l\in\mathcal{S}} w_l \, \ell_{\mathrm{rca}}^l.
\label{eq:rca_total}
\end{equation}
Here, $\mathcal{S}$ denotes the set of selected layers, $\mathcal{P}$ indexes all unordered sample pairs, and $w_l$ controls the contribution of each layer. Unlike BN matching, which only constrains marginal activation statistics, RCA preserves the relative arrangement of samples in feature space. As a result, teacher and student are encouraged to agree not only on \emph{what} activations are produced on average, but also on \emph{how} synthetic samples are organized with respect to one another.

\begin{algorithm}[t]
\small
\caption{STARS-enhanced surrogate synthesis for ANN-to-SNN DFKD.}
\label{alg:star_pipeline}
\begin{algorithmic}[1]
\Require Teacher $f_T$ with BN statistics $\{\hat{\mu}_l,\hat{\sigma}_l^2\}$, SNN student $f_S$, selected layers $\mathcal{S}$, synthesis steps $K$, simulation steps $T$, threshold count $M$, weights $\lambda_{\mathrm{BN}},\lambda_{\mathrm{reg}},\lambda_{\mathrm{rca}},\lambda_{\mathrm{tar}}$
\Ensure Final synthetic batch $\mathcal{B}^{\star}$ for downstream ANN-to-SNN distillation
\State \textbf{Init:} Randomly initialize the synthetic batch $\mathcal{B}$
\algstage{Synthesis stage (optimize surrogate data under teacher priors and STARS constraints)}
\For{$k=1,\ldots,K$}
\State Forward $\mathcal{B}$ through $f_T$ and compute $\mathcal{L}_{\mathrm{cls}}(\mathcal{B};f_T)$ together with the BN regularizer $\mathcal{L}_{\mathrm{BN}}(\mathcal{B})$ \eqcomment{eq:lbn}
\State Forward $\mathcal{B}$ through student $f_S$ for $T$ steps and collect pre-LIF student features $\{F_S^l\}_{l\in\mathcal{S}}$
\State Collect the corresponding teacher features $\{F_T^l\}_{l\in\mathcal{S}}$
\algstage{Layerwise STARS regularization at pre-LIF features}
\For{each $l\in\mathcal{S}$}
\State Time-average $F_S^l$ and apply global average pooling to obtain $z_T^l, z_S^l$ \eqcomment{eq:method_features}
\State \textbf{RCA:} Row-normalize $z_T^l$ and $z_S^l$, form $G_T^l$ and $G_S^l$, and compute $\ell_{\mathrm{rca}}^l$ \eqcomment{eq:rca_layer}
\State Construct teacher-anchored thresholds $\{\theta_m^l\}_{m=1}^{M}$ from $z_T^l$
\State \textbf{TAR:} Compute soft exceedance probabilities $p_T^{l,m}$ and $p_S^{l,m}$, then compute $\ell_{\mathrm{tar}}^l$ \eqcomment{eq:tar_layer}
\EndFor
\State Aggregate layer losses into $\mathcal{L}_{\mathrm{RCA}}$ and $\mathcal{L}_{\mathrm{TAR}}$ \eqcomment{eq:rca_total}
\State $\mathcal{L}_{\mathrm{DFKD}} \leftarrow \mathcal{L}_{\mathrm{cls}}+\lambda_{\mathrm{BN}}\mathcal{L}_{\mathrm{BN}}+\lambda_{\mathrm{reg}}\mathcal{R}$ \eqcomment{eq:dfkd}; $\mathcal{L}_{\mathrm{total}} \leftarrow \mathcal{L}_{\mathrm{DFKD}}+\lambda_{\mathrm{rca}}\mathcal{L}_{\mathrm{RCA}}+\lambda_{\mathrm{tar}}\mathcal{L}_{\mathrm{TAR}}$ \eqcomment{eq:method_overall}
\State Update $\mathcal{B}\leftarrow \mathcal{B}-\eta\nabla_{\mathcal{B}}\mathcal{L}_{\mathrm{total}}$ \Comment{update synthetic batch only}
\EndFor
\State \Return $\mathcal{B}^{\star}\leftarrow\mathcal{B}$
\end{algorithmic}
\end{algorithm}

\textbf{Tail-Aware Regularization (TAR).}
While RCA regularizes relational consistency across samples, it does not yet target the threshold-sensitive mechanism isolated in Sec.~\ref{sec:analysis}. The analysis there shows that the SNN response depends on the amount of probability mass above effective thresholds rather than on first and second order statistics alone. TAR is designed to directly encode this observation. Instead of matching only moments, TAR explicitly regularizes \emph{tail probabilities across multiple teacher-anchored thresholds}, providing a smooth proxy for threshold-crossing behavior.

For each selected layer, we construct a threshold set $\{\theta_m^l\}_{m=1}^{M}$ from the teacher feature distribution, for example from fixed values or teacher quantiles. When normalization is used, both teacher and student features are normalized using teacher-derived statistics before thresholding, so that the thresholds are defined in a stable teacher-centric coordinate system. We then compute the soft exceedance probabilities
\begin{equation}\small
p_T^{l,m}
=
\frac{1}{BC}\sum_{b,c}
\sigma\!\left(\frac{z_{T,bc}^l-\theta_m^l}{\delta}\right),
\qquad
p_S^{l,m}
=
\frac{1}{BC}\sum_{b,c}
\sigma\!\left(\frac{z_{S,bc}^l-\theta_m^l}{\delta}\right),
\label{eq:tar_prob}
\end{equation}
where $\sigma(\cdot)$ is the sigmoid and $\delta>0$ controls the softness of the threshold transition. Each quantity above can be interpreted as a differentiable estimate of the probability mass lying above threshold $\theta_m^l$. The layer-wise regularization term $\ell_{\mathrm{tar}}^l$ and the final tail-aware regularization objective $\mathcal{L}_{\mathrm{TAR}}$ are
\begin{equation}\small
\ell_{\mathrm{tar}}^l
=
\frac{1}{M}\sum_{m=1}^{M}
\left(p_S^{l,m} - p_T^{l,m}\right)^2,\quad\mathcal{L}_{\mathrm{TAR}}
=
\frac{1}{\sum_{l\in\mathcal{S}} w_l}
\sum_{l\in\mathcal{S}} w_l \, \ell_{\mathrm{tar}}^l.
\label{eq:tar_layer}
\end{equation}
This objective is consistent with the theory in Sec.~\ref{sec:analysis}. Eq.~\eqref{eq:rate_analysis} expresses the expected SNN response through exceedance probabilities above threshold sets, and Eq.~\eqref{eq:rate_gap_analysis} shows that mismatched threshold-wise tail mass can produce different SNN responses even under identical BN statistics. TAR turns this observation into a trainable regularizer by explicitly penalizing mismatches in a finite collection of teacher-anchored tail probabilities. As a result, the synthesized features are constrained not only by mean and variance as in BN matching, but also by threshold-relevant tail probabilities, which directly reduces the under-constrained ambiguity identified above.

\textbf{Overall objective.}
Combining the standard teacher-side DFKD objective with the two STARS modules above, we synthesize the surrogate batch by solving
\begin{equation}\small
\mathcal{B}^{\star}
=
\arg\min_{\mathcal{B}}
\Big[
\mathcal{L}_{\mathrm{DFKD}}(\mathcal{B};f_T)
+
\lambda_{\mathrm{rca}}\cdot\mathcal{L}_{\mathrm{RCA}}(\mathcal{B};f_T,f_S)
+
\lambda_{\mathrm{tar}}\cdot\mathcal{L}_{\mathrm{TAR}}(\mathcal{B};f_T,f_S)
\Big].
\label{eq:method_overall}
\end{equation}
Here, $\mathcal{L}_{\mathrm{DFKD}}$ preserves the original teacher-valid synthesis priors, including BN matching and teacher confidence regularization; $\mathcal{L}_{\mathrm{RCA}}$ enforces the cross-sample relational consistency that these priors do not control; and $\mathcal{L}_{\mathrm{TAR}}$ constrains the threshold-relevant tail behavior that is directly tied to LIF firing. The resulting objective keeps the optimization anchored to the ANN teacher while making the synthesized data substantially more informative for the SNN student. In this sense, STARS is not a replacement for conventional DFKD priors, but a targeted augmentation derived from the mismatch identified by the analysis: RCA addresses the missing relational consistency, and TAR addresses the missing threshold-sensitive tail constraints. After obtaining the optimized synthetic batch $\mathcal{B}^{\star}$, we then distill the student according to Eq.~\eqref{eq:kd}, which can be written as
\begin{equation}\small
\mathcal{L}_{\mathrm{KD}}(\mathcal{B}^{\star})
=
\frac{1}{|\mathcal{B}^{\star}|}\sum_{x\in\mathcal{B}^{\star}}
\left[
\mathrm{KL}\!\left(
\sigma\!\left(f_T(x)/\gamma\right)
\mid
\sigma\!\left(f_S(x)/\gamma\right)
\right)
\right].
\end{equation}

\vspace{-2mm}
\section{Experiments}
\label{sec:experiments}
\vspace{-2mm}
\textbf{Experimental setup.}
We evaluate STARS on CIFAR-10 and CIFAR-100~\citep{krizhevsky2009learning} using VGG-~\citep{simonyan2015very} and ResNet-based~\citep{he2016deep} ANN teacher / SNN student pairs, and report top-1 accuracy. We further test on Tiny-ImageNet~\citep{le2015tiny}, a larger benchmark with 200 classes and $64\times64$ images. All SNN students use direct coding and are evaluated with $T=4$ simulation steps, while reported KD baselines follow their original settings. For fair comparison, DeepInversion and CMI are implemented under the efficient DFKD training protocol of FastDFKD~\citep{fastdfkd}. STARS is plugged into each BN-guided synthesis step by adding RCA and TAR to the original BN-matching objective, while keeping the baseline teacher-side priors and distillation schedules unchanged. We set $\lambda_{\mathrm{rca}}=\lambda_{\mathrm{tar}}=1$ and fix the TAR softness parameter to $\delta=0.5$ unless otherwise specified. Further details are provided in Appendix~\ref{app:exp_settings}.

\textbf{Baselines and plug-and-play protocol.}
We compare against ANN-to-SNN KD methods (Joint A-SNN~\citep{guo2023joint}, BKDSNN~\citep{bkdsnn}, LaSNN~\citep{lasnn}, KDSNN~\citep{KDSNN}) and BN-guided DFKD baselines (DeepInversion~\citep{yin2020dreaming}, CMI~\citep{cmi}, NAYER~\citep{nayer}). For each DFKD method $\mathcal{M}$, the plug-and-play variant $\mathcal{M}$\starplug{} augments its synthesis objective with RCA and TAR while leaving teacher-side priors unchanged, isolating STARS's contribution.

\subsection{Experimental Results}
\textbf{Results on CIFAR-10 \& CIFAR-100.}
Tables~\ref{tab:main_results} and~\ref{tab:imagenet_results} show that STARS consistently improves all three DFKD baselines across every model pair and dataset. Gains are most pronounced on NAYER, with STARS yielding +3.5--4.6\% on CIFAR-10 and +4.5--6.7\% on CIFAR-100. NAYER\starplug{} reaches 95.73/78.42 under ResNet-19$\to$S-ResNet-19. On DeepInversion and CMI, \textbf{\textcolor{starblue}{\textsc{STARS}}} also improve by +2--4\% uniformly, confirming that the relational and tail-aware constraints are complementary to diverse synthesis objectives. 
Notably, NAYER\starplug{} on both CIFAR-10 and CIFAR-100 surpasses several data-available KD baselines and nearly closes the gap to the ANN teacher (95.73 vs.\ 96.30 for CIFAR-10; 78.42 vs.\ 80.48 for CIFAR-100), suggesting that structured synthesis constraints can largely substitute for real training data in ANN-to-SNN distillation.

\begin{table*}[t]
\centering
\small
\setlength{\tabcolsep}{4pt}
\renewcommand{\arraystretch}{1.0}
\caption{Summary results on CIFAR-10 and CIFAR-100.
Models prefixed with ``S-'' denote spiking neural networks (SNNs), while unprefixed models are artificial neural networks (ANNs) by default.
``--'' denotes not reported. Teacher accuracy is reported in the order of CIFAR-10 / CIFAR-100.}
\label{tab:main_results}
\vspace{-2mm}
\resizebox{\textwidth}{!}{%
\begin{tabular}{c l l c l c c c}
\toprule
Type & Method & Teacher Model & Teacher Acc. & Student Model & $T$ & CIFAR-10 & CIFAR-100 \\
\midrule

\multirow{7}{*}{KD}
& LaSNN~\citep{lasnn}
    & VGG-16
    & -- / --
    & S-VGG-16
    & 100
    & 91.22 & 61.52 \\
\pairrule

& \multirow{2}{*}{Joint A-SNN~\citep{guo2023joint}}
    & VGG-16
    & -- / --
    & S-VGG-16
    & 1
    & 93.79 & 74.24 \\
&   & VGG-16
    & -- / --
    & S-VGG-16
    & 4
    & -- & -- \\
\pairrule

& \multirow{2}{*}{KDSNN~\citep{KDSNN}}
    & ResNet-19
    & 95.60 / 79.78
    & S-ResNet-19
    & 4
    & 94.36 & 74.08 \\
&   & ResNet-20
    & 91.77 / 68.40
    & S-ResNet-20
    & 4
    & 89.03 & 60.18 \\
\pairrule

& \multirow{2}{*}{BKDSNN~\citep{bkdsnn}}
    & ResNet-19
    & 95.60 / 79.78
    & S-ResNet-19
    & 4
    & 94.64 & 74.95 \\
&   & ResNet-20
    & 91.77 / 68.40
    & S-ResNet-20
    & 4
    & 89.29 & 60.92 \\



\midrule

\multirow{18}{*}{DFKD}


& \multirow{3}{*}{DeepInversion~\citep{yin2020dreaming}}
    & VGG-16
    & 94.32 / 75.39
    & S-VGG-16
    & 4
    & 66.88 & 35.68 \\
&   & ResNet-34
    & 95.70 / 77.94
    & S-ResNet-18
    & 4
    & 73.84 & 41.71 \\
&   & ResNet-19
    & 96.30 / 80.48
    & S-ResNet-19
    & 4
    & 78.05 & 42.25 \\
\cmidrule(lr){2-8}

& \multirow{3}{*}{DeepInversion\starplug}
    & VGG-16
    & 94.32 / 75.39
    & S-VGG-16
    & 4
    & \textbf{70.12} & \textbf{39.72} \\
& 
    & ResNet-34
    & 95.70 / 77.94
    & S-ResNet-18
    & 4
    & \textbf{76.93} & \textbf{46.18} \\
& 
    & ResNet-19
    & 96.30 / 80.48
    & S-ResNet-19
    & 4
    & \textbf{80.97} & \textbf{46.73} \\
\pairrule

& \multirow{3}{*}{CMI~\citep{cmi}}
    & VGG-16
    & 94.32 / 75.39
    & S-VGG-16
    & 4
    & 80.20 & 55.91 \\
&   & ResNet-34
    & 95.70 / 77.94
    & S-ResNet-18
    & 4
    & 83.13 & 58.02 \\
&   & ResNet-19
    & 96.30 / 80.48
    & S-ResNet-19
    & 4
    & 85.44 & 60.72 \\
\cmidrule(lr){2-8}

& \multirow{3}{*}{CMI\starplug}
    & VGG-16
    & 94.32 / 75.39
    & S-VGG-16
    & 4
    & \textbf{83.67} & \textbf{58.96} \\
& 
    & ResNet-34
    & 95.70 / 77.94
    & S-ResNet-18
    & 4
    & \textbf{85.85} & \textbf{60.28} \\
& 
    & ResNet-19
    & 96.30 / 80.48
    & S-ResNet-19
    & 4
    & \textbf{88.02} & \textbf{63.49} \\
\pairrule

& \multirow{3}{*}{NAYER~\citep{nayer}}
    & VGG-16
    & 94.32 / 75.39
    & S-VGG-16
    & 4
    & 88.30 & 68.09 \\
&   & ResNet-34
    & 95.70 / 77.94
    & S-ResNet-18
    & 4
    & 90.54 & 71.89 \\
&   & ResNet-19
    & 96.30 / 80.48
    & S-ResNet-19
    & 4
    & 92.02 & 73.95 \\
\cmidrule(lr){2-8}

& \multirow{3}{*}{NAYER\starplug}
    & VGG-16
    & 94.32 / 75.39
    & S-VGG-16
    & 4
    & \textbf{92.89}  & \textbf{74.79} \\
& 
    & ResNet-34
    & 95.70 / 77.94
    & S-ResNet-18
    & 4
    & \textbf{94.08} & \textbf{76.97} \\
& 
    & ResNet-19
    & 96.30 / 80.48
    & S-ResNet-19
    & 4
    & \textbf{95.73} & \textbf{78.42} \\

\bottomrule
\end{tabular}%
}
\vspace{-4mm}
\end{table*}

\begin{table*}[t]
\centering
\small
\setlength{\tabcolsep}{4pt}
\renewcommand{\arraystretch}{1.0}
\caption{Tiny-ImageNet DFKD results under the ResNet-34 $\rightarrow$ S-ResNet-18 setting. Tiny-ImageNet contains 200 classes with $64\times 64$ images. $\Delta$ denotes the Top-1 Acc. improvement of each STARS-enhanced variant over its corresponding DFKD baseline.}
\label{tab:imagenet_results}
\vspace{-2mm}
\resizebox{\textwidth}{!}{%
\begin{tabular}{c l l c l c c c}
\toprule
Type & Method & Teacher Model & Teacher Acc. & Student Model & $T$ & Top-1 Acc. & $\Delta$ \\
\midrule
\multirow{4}{*}{DFKD}
& CMI
    & ResNet-34
    & 67.83
    & S-ResNet-18
    & 4
    & 57.51
    & -- \\
\cmidrule(lr){2-8}

&
    CMI\starplug
    & ResNet-34
    & 67.83
    & S-ResNet-18
    & 4
    & \textbf{60.84}
    & \textbf{+3.33} \\
\cmidrule(lr){2-8}

&
    NAYER
    & ResNet-34
    & 67.83
    & S-ResNet-18
    & 4
    & 61.21
    & -- \\
\cmidrule(lr){2-8}

&
    NAYER\starplug
    & ResNet-34
    & 67.83
    & S-ResNet-18
    & 4
    & \textbf{64.18}
    & \textbf{+2.97} \\

\bottomrule
\end{tabular}%
}
\vspace{-4mm}
\end{table*}

\textbf{Results on Tiny-ImageNet.} 
On Tiny-ImageNet, improvements remain consistent, demonstrating generalization beyond CIFAR scales.
Specifically, under the same ResNet-34 $\rightarrow$ S-ResNet-18 setting, \textbf{\textcolor{starblue}{\textsc{STARS}}} improves from 57.51 to 60.84 on CMI, and from 61.21 to 64.18 on NAYER, corresponding to gains of 3.33 and 2.97, respectively. These results indicate that the benefits of relational consistency and tail-aware regularization are not limited to CIFAR-scale benchmarks, but remain useful when the synthesis problem becomes harder due to a larger class space and more complex image distribution.


\begin{wraptable}{r}{0.5\textwidth}
\vspace{-5mm}
\centering
\small
\caption{Component ablation of STARS on CIFAR-100 under the ResNet-19 $\rightarrow$ S-ResNet-19 setting.}
\vspace{-2mm}
\label{tab:star_ablation}
\setlength{\tabcolsep}{3.5pt}
\renewcommand{\arraystretch}{1.1}
\begin{tabular}{l||ccc|c}
\toprule
Variant & BN-only & RCA only  & TAR only & STARS\\
\midrule
Acc (\%)& 73.95 & 76.18 & 74.36 & \textbf{78.42}\\
\bottomrule
\end{tabular}
\vspace{-4mm}
\end{wraptable}
\textbf{Component ablation of STARS.} Table~\ref{tab:star_ablation} isolates the effects of the two components in \textbf{\textcolor{starblue}{\textsc{STARS}}}. This experiment focuses on the ResNet-19 $\rightarrow$ S-ResNet-19 setting on CIFAR-100, where we examine whether RCA and TAR are individually useful and whether their combination yields complementary gains. The ablation shows that both components contribute positively, while the full STARS objective yields the strongest result. In particular, adding RCA alone improves the BN-only baseline from 73.95 to 76.18, whereas adding TAR alone gives a smaller but still consistent gain to 74.36. Combining both modules reaches 78.42, indicating that relational consistency and tail-aware regularization address complementary aspects of the synthesis mismatch.

%

\textbf{Tail-quantile ablation for TAR.} Table~\ref{tab:star_hparam} reports an ablation of the tail quantile set used in TAR on CIFAR-100 under ResNet-19 $\rightarrow$ S-ResNet-19. 
Since TAR matches teacher and student soft
\begin{wraptable}{r}{0.4\textwidth}
\centering
\small
\caption{Ablation of the tail quantile set used in TAR on CIFAR-100 under the ResNet-19 $\rightarrow$ S-ResNet-19 setting. $M$ is the number of teacher-anchored thresholds, \textit{i.e.,} the length of the quantile set.}
\vspace{-2mm}
\label{tab:star_hparam}
\setlength{\tabcolsep}{1.5pt}
\renewcommand{\arraystretch}{0.8}
\begin{tabular}{l c c}
\toprule
Tail quantiles & $M$ & Acc. (\%) \\
\midrule
$(0.90, 0.95)$ & 2 & 76.58 \\
$(0.80, 0.90, 0.95)$ & 3 & 77.46 \\
$(0.70, 0.80, 0.90, 0.95)$ & 4 & 78.07 \\
$(0.60, 0.70, 0.80, 0.90, 0.95)$ & 5 & \textbf{78.42} \\
\bottomrule
\end{tabular}
\vspace{-2mm}
\end{wraptable}
exceedance probabilities over teacher-anchored thresholds, 
this experiment tests 
how much threshold-wise tail coverage is needed for effective ANN-to-SNN transfer. As the number of matched quantiles increases from $M=2$ to $M=5$, the accuracy improves monotonically from 76.58 to 77.46, 78.07, and finally 78.42. This trend is consistent with the theoretical motivation of STARS: richer coverage of spike-relevant tail regions provides a better approximation to the threshold-sensitive response statistics that BN matching alone does not determine. At least within the tested range, the results indicate that matching a broader set of informative tail thresholds yields more effective synthetic supervision for the SNN student.

\begin{figure}[t]
    \centering
    \includegraphics[width=\linewidth]{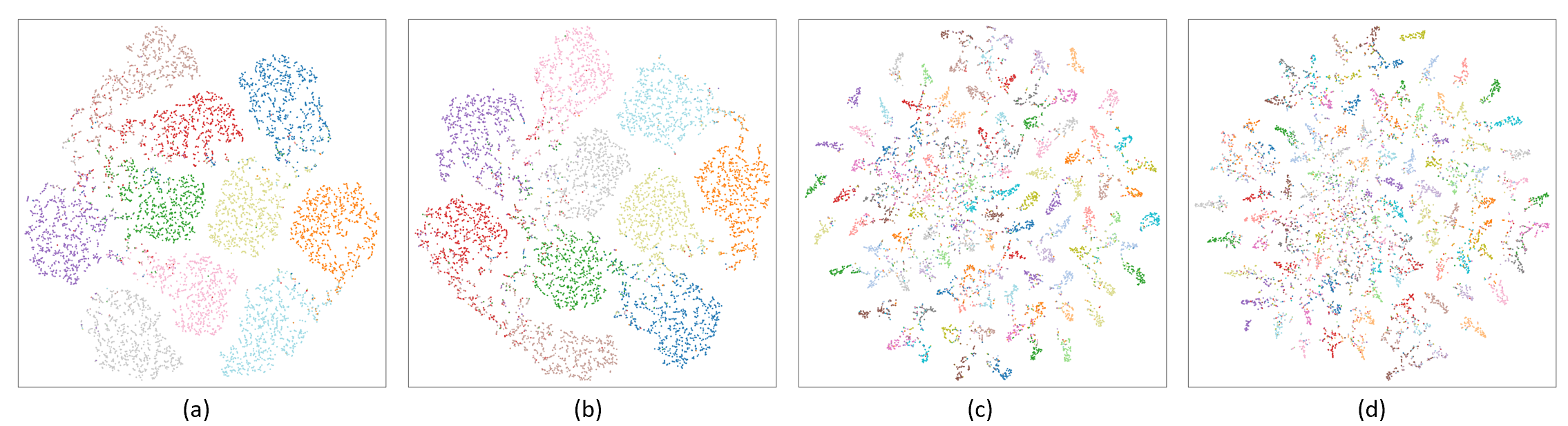}
    \vspace{-5mm}
    \caption{t-SNE visualization of SNN student (S-VGG-16) penultimate-layer features under NAYER-based DFKD (VGG-16$\to$S-VGG-16, $T=4$). Each color denotes one class. \textbf{(a)} CIFAR-10 with STARS, \textbf{(b)} CIFAR-10 with NAYER only, \textbf{(c)} CIFAR-100 with STARS, \textbf{(d)} CIFAR-100 with NAYER only. In \textbf{(a)} and \textbf{(c)}, class clusters are noticeably tighter and more linearly separable with well-defined inter-class margins, whereas \textbf{(b)} and \textbf{(d)} exhibit considerably more diffuse and entangled boundaries, showing that STARS produces richer and more discriminative training signals for the SNN student.}
    \label{fig:tsne_vis}
    \vspace{-3mm}
\end{figure}

\textbf{Visualization.} Figure~\ref{fig:tsne_vis} shows t-SNE visualizations~\citep{maaten2008visualizing} of the penultimate-layer features of the SNN student (S-VGG-16) under the VGG-16$\to$S-VGG-16 setting with $T=4$, using NAYER as the base DFKD method. From left to right, the panels show: (a) CIFAR-10 with STARS, (b) CIFAR-10 with NAYER only, (c) CIFAR-100 with STARS, and (d) CIFAR-100 with NAYER only. Compared with NAYER alone, \textbf{\textcolor{starblue}{\textsc{STARS}}} leads to a clearer tendency toward more compact intra-class grouping and reduced inter-class entanglement on both datasets. This qualitative trend is consistent with the quantitative gains in the main results, suggesting that STARS provides more informative synthetic supervision for learning discriminative SNN representations.

\textbf{Impact of time steps $T$.}
Appendix Table~\ref{tab:app_time_steps} evaluates \textbf{\textcolor{starblue}{\textsc{STARS}}} under different simulation time steps $T$. The consistent gains show that STARS remains effective for SNN students with different temporal resolutions, further demonstrating its plug-and-play applicability.

\textbf{Impact of neuron models.}
Appendix Table~\ref{tab:app_neuron_models} further tests different spiking neuron models. \textbf{\textcolor{starblue}{\textsc{STARS}}} consistently improves the base DFKD method across these variants, indicating that the proposed relational and tail-aware constraints are not tied to a single neuron implementation.

\vspace{-2mm}
\section{Conclusion}
\label{sec:conclusion}
\vspace{-2mm}
We presented STARS, a plug-and-play framework for ANN-to-SNN data-free knowledge distillation that addresses the mismatch between ANN-oriented BN-guided synthesis and SNN student dynamics. Through a formal analysis of the BN-moment ambiguity, we showed that matching first- and second-order ANN statistics leaves spike responses under-constrained. STARS resolves this by augmenting standard synthesis with Relational Consistency Alignment (RCA), which preserves cross-sample relational structure between teacher and student, and Tail-Aware Regularization (TAR), which constrains threshold-relevant tail probabilities through soft exceedance. Extensive experiments on CIFAR-10, CIFAR-100, and Tiny-ImageNet demonstrate that STARS consistently improves three diverse DFKD baselines across multiple ANN-SNN teacher-student pairs, with NAYER+STARS achieving gains of up to 4.6\% on CIFAR-10 and 6.7\% on CIFAR-100 over the baseline. These results establish that complementing BN matching with relational and tail-aware constraints is an effective and generalizable strategy for SNN-oriented data-free distillation. 

\newpage

\bibliographystyle{unsrt}
\bibliography{ref}

\newpage
\appendix

\section{Detailed Experimental Settings}
\label{app:exp_settings}

All experiments are conducted on NVIDIA RTX Pro 6000 GPUs. This appendix summarizes the detailed experimental settings used in our DFKD experiments. Unless otherwise specified, STARS is added on top of the base synthesis objective with $\lambda_{\mathrm{rca}} = \lambda_{\mathrm{tar}} = 1$, \texttt{threshold\_mode=quantile}, and \texttt{threshold\_values=$\{$0.6,0.7,0.8,0.9,0.95$\}$}.

\paragraph{DeepInversion.}
For DeepInversion on CIFAR-10 and CIFAR-100, we evaluate across three ANN teacher / SNN student pairs: VGG-16 $\rightarrow$ S-VGG-16, ResNet-34 $\rightarrow$ S-ResNet-18, and ResNet-19 $\rightarrow$ S-ResNet-19, all with $T=4$ simulation steps. The following training hyperparameters are shared across all three pairs. The student is trained for 200 epochs with no warmup. We use 400 KD steps, 400 EP steps, and 1000 generator steps, with student batch size 256 and synthesis batch size 256. The student learning rate is 0.1 and the generator learning rate is 0.1. The DeepInversion loss weights are set to adversarial loss 1.0, BN regularization 10.0, and one-hot loss 1.0, and the KD temperature is 20. The random seed is fixed to 0. On top of this baseline, STARS applies RCA and TAR with \texttt{threshold\_values=$\{$0.6,0.7,0.8,0.9,0.95$\}$}, shared across all three pairs.

\paragraph{CMI.}
For CMI on CIFAR-10 and CIFAR-100, we likewise evaluate across the same three ANN teacher / SNN student pairs: VGG-16 $\rightarrow$ S-VGG-16, ResNet-34 $\rightarrow$ S-ResNet-18, and ResNet-19 $\rightarrow$ S-ResNet-19, all with $T=4$. The following training hyperparameters are shared across all three pairs. The student is trained for 200 epochs without warmup, and the inversion / distillation schedule uses 400 KD steps, 400 EP steps, and 200 generator steps. Both the student batch size and the synthesis batch size are 256. The student learning rate is 0.1 and the generator learning rate is $10^{-3}$. The CMI objective uses loss weights 0.5 for adversarial loss, 1.0 for BN regularization, 1.0 for one-hot loss, 0.8 for contrastive regularization, and 0.2 for teacher-side contrastive temperature scaling, with KD temperature 20 and seed 0. The STARS configuration is identical to that used in DeepInversion above and is likewise shared across all three pairs. For Tiny-ImageNet, CMI also follows this same training and synthesis configuration under the ResNet-34 $\rightarrow$ S-ResNet-18 setting with $T=4$.

\paragraph{NAYER.}
For NAYER on CIFAR-10 and CIFAR-100, we evaluate across three ANN teacher / SNN student pairs: VGG-16 $\rightarrow$ S-VGG-16, ResNet-34 $\rightarrow$ S-ResNet-18, and ResNet-19 $\rightarrow$ S-ResNet-19, all with $T=4$ simulation steps. The student is trained for 320 epochs with 20 warmup epochs. The student learning rate is 0.2 and the generator learning rate is $4\times 10^{-3}$. The NAYER loss weights are adversarial loss 1.33, BN regularization 20.0, and one-hot loss 0.75. We use 1000 KD steps, 1000 EP steps, and 75 generator steps per synthesis round, with generator life 10, generator loops 4, and GWP loops 10. The random seed is fixed to 0.

For Tiny-ImageNet, we use the ResNet-34 $\rightarrow$ S-ResNet-18 setting with $T=4$ and input resolution $64\times 64$. The teacher top-1 accuracy is 67.83. The student is trained for 320 epochs with 20 warmup epochs. We use 1000 KD steps, 1000 EP steps, and 60 generator steps, together with generator life 10, generator loops 4, and GWP loops 20. The student batch size is 256 and the synthesis batch size is 200. The student learning rate is 0.1 and the generator learning rate is $4\times 10^{-3}$. The NAYER loss weights are adversarial loss 1.33, BN regularization 10.0, and one-hot loss 0.5, with seed 0.

\section{Supplementary Discussion on BN-Moment Ambiguity in SNNs}
\label{app:bn_ambiguity}

In this appendix, we provide supplementary remarks for the simplified single-layer direct-coding analysis in Sec.~\ref{sec:analysis}. Our goal is not to establish a fully general theorem for deep SNNs, but to further clarify, step by step, why matching ANN-side BN moments is insufficient to uniquely characterize the spike response of an SNN student.

\subsection{BN moments do not uniquely determine spike responses}

Recall that under the simplified direct-coding LIF approximation in Sec.\ref{sec:analysis}, the average spike response over $T$ time steps is
\begin{equation}\small
\bar S_T(a)=\frac{1}{T}\sum_{t=1}^{T}\mathbf{1}(a\ge \theta_t),
\label{eq:appendix_avg_spike}
\end{equation}
and the corresponding expected firing rate under an activation distribution $P$ is
\begin{equation}\small
R_T(P)
=
\mathbb{E}_{a\sim P}\!\left[\bar S_T(a)\right]
=
\frac{1}{T}\sum_{t=1}^{T}\bigl(1-F_P(\theta_t)\bigr),
\label{eq:appendix_rate}
\end{equation}
where $F_P$ is the cumulative distribution function (CDF) of $P$.

\begin{remark}
Under the simplified single-layer direct-coding LIF approximation, matching the first and second order statistics of the ANN-side pre-BN activation distribution is generally insufficient to uniquely determine the expected spike response of the corresponding SNN neuron.
\end{remark}

\textbf{Explanation.}
We establish this claim through four explicit steps.

\textbf{Step 1: The expected firing rate is a linear functional of tail probabilities.}
By the linearity of expectation and the indicator representation in Eq.~\eqref{eq:appendix_avg_spike},
\begin{equation}\small
R_T(P)
= \frac{1}{T}\sum_{t=1}^{T}\mathbb{E}_{a\sim P}\bigl[\mathbf{1}(a\ge\theta_t)\bigr]
= \frac{1}{T}\sum_{t=1}^{T}\mathbb{P}_{a\sim P}(a\ge\theta_t)
= \frac{1}{T}\sum_{t=1}^{T}\bigl(1-F_P(\theta_t)\bigr).
\label{eq:appendix_rate_expand}
\end{equation}
Hence $R_T(P)$ is fully determined by the right-tail survival probabilities $\{1-F_P(\theta_t)\}_{t=1}^{T}$, i.e., the probability masses that the distribution $P$ places above each effective threshold.

\textbf{Step 2: BN matching constrains only the first two moments.}
The BN-based synthesis objective enforces that the synthetic distribution matches the stored BN statistics of the teacher, which amounts to matching the mean and variance of the pre-BN activation:
\begin{equation}\small
\mu_A(P) = \mathbb{E}_{a\sim P}[a], \qquad \sigma_A^2(P) = \mathrm{Var}_{a\sim P}(a).
\label{eq:appendix_bn_moments}
\end{equation}
No constraint is imposed on higher-order statistics or on any aspect of the shape of $F_P$ beyond these two quantities.

\textbf{Step 3: The first two moments do not uniquely determine tail probabilities.}
In general, specifying the mean and variance of a distribution does not uniquely determine its CDF, and therefore does not uniquely determine the tail probabilities $\{1-F_P(\theta_t)\}$. To see this concretely, let $(\mu,\sigma^2)$ be any fixed target mean and variance, and let $P_r$ be a reference distribution (\textit{e.g.,} a Gaussian $\mathcal{N}(\mu,\sigma^2)$) with those statistics. For any threshold $\theta$, one can construct a family of distributions that share the same mean and variance as $P_r$ but place strictly different probability mass above $\theta$. For example, consider the mixture
\begin{equation}\small
P_s = (1-\epsilon)\,P_r + \epsilon\,\delta_{\mu + c},
\label{eq:appendix_mixture}
\end{equation}
where $\delta_{\mu+c}$ is a point mass at $\mu+c$ for some shift $c\ne 0$, and $\epsilon>0$ is small. To preserve the variance $\sigma^2$, one may instead mix $P_r$ with a two-point distribution supported at $\mu\pm\Delta$ that has the same mean $\mu$ but different mass distribution above $\theta$. More generally, a standard result in moment theory states that a distribution is not uniquely determined by finitely many moments, so for any target $(\mu,\sigma^2)$ there exist infinitely many distributions with those moments that differ in their higher-order shape and hence in their tail probabilities at $\theta$.

\textbf{Step 4: Consequence for the spike response.}
Combining Steps 1--3: since $R_T(P)$ is determined by tail probabilities at $\{\theta_t\}$ (Step 1), and BN matching leaves those tail probabilities under-constrained (Steps 2--3), two distributions $P$ and $Q$ satisfying
\begin{equation}\small
\mu_A(P)=\mu_A(Q), \qquad \sigma_A^2(P)=\sigma_A^2(Q)
\label{eq:appendix_bn_equiv}
\end{equation}
may nonetheless satisfy $F_P(\theta_t)\ne F_Q(\theta_t)$ for some $t\in\{1,\dots,T\}$, which by Eq.~\eqref{eq:appendix_rate_expand} directly yields
\begin{equation}\small
R_T(P) \ne R_T(Q).
\label{eq:appendix_rate_differ}
\end{equation}
This establishes the claimed insufficiency: BN moment matching cannot, in general, uniquely determine the expected spike response.

\subsection{Operating-point interpretation}

The above ambiguity has a direct interpretation in terms of the operating regime of the SNN neuron. We consider two canonical cases, each derived directly from Eq.~\eqref{eq:appendix_rate}.

\textbf{Under-firing shift.}
Suppose the synthetic distribution $P_s$ has a CDF that dominates the reference distribution $P_r$ stochastically at every threshold, i.e.,
\begin{equation}\small
F_{P_s}(\theta_t)\ge F_{P_r}(\theta_t), \qquad \forall\, t\in\{1,\dots,T\},
\label{eq:appendix_under}
\end{equation}
with strict inequality for at least one $t$. Taking complements gives the corresponding condition on the survival probabilities:
\begin{equation}\small
1 - F_{P_s}(\theta_t) \le 1 - F_{P_r}(\theta_t), \qquad \forall\, t\in\{1,\dots,T\}.
\label{eq:appendix_under_survival}
\end{equation}
Summing over $t=1,\dots,T$, dividing by $T$, and applying Eq.~\eqref{eq:appendix_rate}, we obtain
\begin{equation}\small
R_T(P_s)
= \frac{1}{T}\sum_{t=1}^{T}\bigl(1-F_{P_s}(\theta_t)\bigr)
\;\le\; \frac{1}{T}\sum_{t=1}^{T}\bigl(1-F_{P_r}(\theta_t)\bigr)
= R_T(P_r).
\label{eq:appendix_under_ineq}
\end{equation}
Since the inequality in Eq.~\eqref{eq:appendix_under} is strict for at least one $t$, the corresponding term in the sum contributes a strict inequality, so the overall sum is strictly less:
\begin{equation}\small
R_T(P_s)<R_T(P_r).
\label{eq:appendix_under_rate}
\end{equation}
Intuitively, $P_s$ places less probability mass above the effective thresholds than $P_r$ does, so the neuron fires less frequently under $P_s$ than under $P_r$. This constitutes an \emph{under-firing} shift.

\textbf{Over-firing shift.}
Conversely, suppose
\begin{equation}\small
F_{P_s}(\theta_t)\le F_{P_r}(\theta_t), \qquad \forall\, t\in\{1,\dots,T\},
\label{eq:appendix_over}
\end{equation}
with strict inequality for at least one $t$. Taking complements reverses the inequalities on the survival probabilities:
\begin{equation}\small
1 - F_{P_s}(\theta_t) \ge 1 - F_{P_r}(\theta_t), \qquad \forall\, t\in\{1,\dots,T\}.
\label{eq:appendix_over_survival}
\end{equation}
Summing over $t$ and applying Eq.~\eqref{eq:appendix_rate} gives
\begin{equation}\small
R_T(P_s)
= \frac{1}{T}\sum_{t=1}^{T}\bigl(1-F_{P_s}(\theta_t)\bigr)
\;\ge\; \frac{1}{T}\sum_{t=1}^{T}\bigl(1-F_{P_r}(\theta_t)\bigr)
= R_T(P_r),
\label{eq:appendix_over_ineq}
\end{equation}
and strict inequality for at least one term yields
\begin{equation}\small
R_T(P_s)>R_T(P_r).
\label{eq:appendix_over_rate}
\end{equation}
In this case, $P_s$ places more probability mass above the effective thresholds than $P_r$, causing the neuron to fire more frequently. This constitutes an \emph{over-firing} or early-saturation shift.

\subsection{Implication}

These supplementary observations reinforce the main point of Sec.~\ref{sec:analysis}: ANN-side BN matching defines an equivalence class of activation distributions from the teacher's perspective, but this equivalence does not carry over to the SNN student. Even when two activation distributions share the same mean and variance, they may induce substantially different spike responses because the SNN response is governed by threshold-crossing tail masses rather than by the first and second order statistics alone. In practical terms, this underdetermination creates a degree of freedom that can systematically push synthesized activations into either an under-firing or over-firing regime (as demonstrated in the two cases above), degrading the quality of knowledge transfer from the ANN teacher to the SNN student. This is precisely why constraining the threshold-relevant tail behavior---as done by the TAR component of STARS---is a principled and necessary complement to the standard BN-prior-based synthesis objective.

\section{Additional Robustness Studies}
\label{app:robustness}

\begin{table}[h]
\centering
\small
\caption{Impact of simulation time steps $T$ on CIFAR-100 under the VGG-16 $\rightarrow$ S-VGG-16 setting. We report Top-1 Acc. (\%) of NAYER\starplug{} with different SNN student simulation steps.}
\label{tab:app_time_steps}
\setlength{\tabcolsep}{8pt}
\renewcommand{\arraystretch}{1.05}
\begin{tabular}{l l c c}
\toprule
Teacher & Student & $T$ & Top-1 Acc. (\%) \\
\midrule
VGG-16 & S-VGG-16 & 2 & 72.48 \\
VGG-16 & S-VGG-16 & 4 & 74.79 \\
VGG-16 & S-VGG-16 & 8 & 75.25 \\
\bottomrule
\end{tabular}
\end{table}

\begin{table}[h]
\centering
\small
\caption{Impact of different spiking neuron models on CIFAR-100 under the VGG-16 $\rightarrow$ S-VGG-16 setting with $T=4$. We report Top-1 Acc. (\%) of NAYER\starplug{} when replacing the student neuron model.}
\label{tab:app_neuron_models}
\setlength{\tabcolsep}{8pt}
\renewcommand{\arraystretch}{1.05}
\begin{tabular}{l l c c}
\toprule
Teacher & Student & Neuron model & Top-1 Acc. (\%) \\
\midrule
VGG-16 & S-VGG-16 & IF & 74.88 \\
VGG-16 & S-VGG-16 & LIF & 74.79 \\
VGG-16 & S-VGG-16 & PLIF & 75.21 \\
\bottomrule
\end{tabular}
\end{table}

\section{Limitations}
\label{app:limitations}

Since conventional Vision Transformers typically use LayerNorm rather than BatchNorm, the current BN-guided DFKD setting cannot be directly applied to transformer-based SNNs. Extending STARS beyond BN-guided synthesis to transformer-based SNN architectures is therefore left as an important direction for future work.

\section{Broader Impact}
\label{app:impact}

This work advances ANN-to-SNN data-free knowledge distillation, a technique that enables training energy-efficient spiking neural networks without access to the original training data. By removing the dependency on real data during distillation, STARS reduces barriers related to data privacy, proprietary datasets, and storage constraints, which may facilitate the deployment of efficient SNN models in privacy-sensitive applications such as on-device medical sensing and edge robotics.

From a scientific standpoint, the BN-moment ambiguity analysis reveals a fundamental gap between ANN-oriented data synthesis and SNN student requirements. We hope this analysis encourages the community to think more carefully about the mismatch between synthesis objectives and student architectures in heterogeneous knowledge distillation settings.

We do not foresee direct negative societal impacts specific to this work. SNNs are primarily studied for their energy efficiency and biological plausibility, and the distillation framework proposed here does not enable any new harmful capability. As with any machine learning method, the underlying models could in principle be used in applications with ethical implications, but those risks are no greater than those associated with standard neural network training.
\newpage
\input{checklist.tex}

\end{document}

%% file: checklist.tex
\section*{NeurIPS Paper Checklist}

\begin{enumerate}

\item {\bf Claims}
    \item[] Question: Do the main claims made in the abstract and introduction accurately reflect the paper's contributions and scope?
    \item[] Answer: \answerYes{}
    \item[] Justification: The abstract and introduction clearly state the three contributions: (1) analysis of BN-moment ambiguity for SNN students (Sec.~\ref{sec:analysis}), (2) the STARS framework with RCA and TAR (Sec.~\ref{sec:method}), and (3) plug-and-play gains on CIFAR-10, CIFAR-100, and Tiny-ImageNet (Sec.~\ref{sec:experiments}). All claims are supported by the corresponding experimental results in Tables~\ref{tab:main_results} and~\ref{tab:imagenet_results}.

\item {\bf Limitations}
    \item[] Question: Does the paper discuss the limitations of the work performed by the authors?
    \item[] Answer: \answerYes{}
    \item[] Justification: Limitations are discussed in Appendix~\ref{app:limitations}, covering computational overhead of the additional student forward pass, the scope of the theoretical analysis (single-layer subthreshold approximation), and the restricted evaluation benchmarks (image classification on CIFAR and Tiny-ImageNet).

\item {\bf Theory assumptions and proofs}
    \item[] Question: For each theoretical result, does the paper provide the full set of assumptions and a complete (and correct) proof?
    \item[] Answer: \answerYes{}
    \item[] Justification: The analysis in Sec.~\ref{sec:analysis} explicitly states all assumptions (direct coding, subthreshold regime, single-layer approximation, reset potential $V_r=0$) and derives all results step by step through Eqs.~\eqref{eq:subthreshold_analysis}--\eqref{eq:rate_gap_analysis}. Extended step-by-step derivations are provided in Appendix~\ref{app:bn_ambiguity}.

\item {\bf Experimental result reproducibility}
    \item[] Question: Does the paper fully disclose all the information needed to reproduce the main experimental results of the paper to the extent that it affects the main claims and/or conclusions of the paper (regardless of whether the code and data are provided or not)?
    \item[] Answer: \answerYes{}
    \item[] Justification: Appendix~\ref{app:exp_settings} provides full hyperparameters for all experiments across all three DFKD baselines (DeepInversion, CMI, NAYER) on all datasets, including learning rates, loss weights, training schedules, batch sizes, and random seeds.

\item {\bf Open access to data and code}
    \item[] Question: Does the paper provide open access to the data and code, with sufficient instructions to faithfully reproduce the main experimental results, as described in supplemental material?
    \item[] Answer: \answerNo{}
    \item[] Justification: Code is not yet publicly released at submission time. All datasets used (CIFAR-10, CIFAR-100, Tiny-ImageNet) are publicly available standard benchmarks. We plan to release the code upon acceptance.

\item {\bf Experimental setting/details}
    \item[] Question: Does the paper specify all the training and test details (e.g., data splits, hyperparameters, how they were chosen, type of optimizer) necessary to understand the results?
    \item[] Answer: \answerYes{}
    \item[] Justification: Sec.~\ref{sec:experiments} describes the overall experimental setup, and Appendix~\ref{app:exp_settings} provides complete per-baseline hyperparameter tables including optimizer settings, training epochs, loss weights, and synthesis schedules.

\item {\bf Experiment statistical significance}
    \item[] Question: Does the paper report error bars suitably and correctly defined or other appropriate information about the statistical significance of the experiments?
    \item[] Answer: \answerNo{}
    \item[] Justification: Results are reported as single-run top-1 accuracy, consistent with the reporting practice of compared baselines in the SNN distillation literature. Error bars are not provided due to the high computational cost of running multiple full training trials for each of the 18 experimental configurations.

\item {\bf Experiments compute resources}
    \item[] Question: For each experiment, does the paper provide sufficient information on the computer resources (type of compute workers, memory, time of execution) needed to reproduce the experiments?
    \item[] Answer: \answerYes{}
    \item[] Justification: Appendix~\ref{app:exp_settings} states that all experiments are conducted on NVIDIA RTX Pro 6000 GPUs. The number of training epochs and synthesis steps are specified per configuration, providing a basis for runtime estimation.

\item {\bf Code of ethics}
    \item[] Question: Does the research conducted in the paper conform, in every respect, with the NeurIPS Code of Ethics \url{https://neurips.cc/public/EthicsGuidelines}?
    \item[] Answer: \answerYes{}
    \item[] Justification: This work involves standard supervised image classification on publicly available benchmarks. No human subjects, sensitive data, or dual-use concerns are involved.

\item {\bf Broader impacts}
    \item[] Question: Does the paper discuss both potential positive societal impacts and negative societal impacts of the work performed?
    \item[] Answer: \answerYes{}
    \item[] Justification: Broader impacts are discussed in Appendix~\ref{app:impact}, covering positive impacts (enabling privacy-preserving and energy-efficient SNN deployment in edge and medical sensing scenarios) and negative societal impacts (none identified beyond general ML misuse risk, which is acknowledged).

\item {\bf Safeguards}
    \item[] Question: Does the paper describe safeguards that have been put in place for responsible release of data or models that have a high risk for misuse (e.g., pre-trained language models, image generators, or scraped datasets)?
    \item[] Answer: \answerNA{}
    \item[] Justification: This paper proposes a distillation regularization framework for image classification SNNs. No generative model, scraped dataset, or high-risk asset is released.

\item {\bf Licenses for existing assets}
    \item[] Question: Are the creators or original owners of assets (e.g., code, data, models), used in the paper, properly credited and are the license and terms of use explicitly mentioned and properly respected?
    \item[] Answer: \answerYes{}
    \item[] Justification: All datasets (CIFAR-10/100~\citep{krizhevsky2009learning}, Tiny-ImageNet~\citep{le2015tiny}) and all baseline methods are cited in the paper. These are standard academic benchmarks and open-source methods distributed for research use.

\item {\bf New assets}
    \item[] Question: Are new assets introduced in the paper well documented and is the documentation provided alongside the assets?
    \item[] Answer: \answerNA{}
    \item[] Justification: This paper does not introduce new datasets, benchmarks, or pre-trained model checkpoints as standalone assets.

\item {\bf Crowdsourcing and research with human subjects}
    \item[] Question: For crowdsourcing experiments and research with human subjects, does the paper include the full text of instructions given to participants and screenshots, if applicable, as well as details about compensation (if any)?
    \item[] Answer: \answerNA{}
    \item[] Justification: This paper does not involve crowdsourcing or research with human subjects.

\item {\bf Institutional review board (IRB) approvals or equivalent for research with human subjects}
    \item[] Question: Does the paper describe potential risks incurred by study participants, whether such risks were disclosed to the subjects, and whether Institutional Review Board (IRB) approvals (or an equivalent approval/review based on the requirements of your country or institution) were obtained?
    \item[] Answer: \answerNA{}
    \item[] Justification: This paper does not involve human subjects research.

\item {\bf Declaration of LLM usage}
    \item[] Question: Does the paper describe the usage of LLMs if it is an important, original, or non-standard component of the core methods in this research? Note that if the LLM is used only for writing, editing, or formatting purposes and does \emph{not} impact the core methodology, scientific rigor, or originality of the research, declaration is not required.
    \item[] Answer: \answerNA{}
    \item[] Justification: LLMs are not used as any component of the proposed method. LLMs were used solely for writing assistance and do not affect the scientific methodology or originality of the research.

\end{enumerate}